\documentclass[conference]{IEEEtran}
\IEEEoverridecommandlockouts
\pdfoutput=1
\usepackage{makecell}
\usepackage{hyperref}
\usepackage{amsmath,amssymb,amsfonts}
\usepackage[linesnumbered,ruled,vlined]{algorithm2e}
\usepackage{graphicx}
\usepackage{textcomp}
\usepackage{xcolor}
\usepackage[numbers,sort&compress]{natbib}
\usepackage{marvosym}

\def\BibTeX{{\rm B\kern-.05em{\sc i\kern-.025em b}\kern-.08em
		T\kern-.1667em\lower.7ex\hbox{E}\kern-.125emX}}
\begin{document}
	
	\title{Evolutionary Generative Adversarial Networks with Crossover Based Knowledge Distillation\\
	}
	\author{\IEEEauthorblockN{Junjie Li,	\thanks{\IEEEauthorrefmark{1} Corresponding author.}
			Junwei Zhang, Xiaoyu Gong,
			Shuai Lü\IEEEauthorrefmark{1}}
		\IEEEauthorblockA{\textit{Key Laboratory of Symbolic Computation and Knowledge Engineering (Jilin University), }\\
			\textit{Ministry of Education, Changchun 130012, China}\\
			\textit{College of Computer Science and Technology, Jilin University, Changchun 130012, China}\\
			\{\href{mailto:junjiel18@mails.jlu.edu.cn}{junjiel18},
			\href{mailto:junwei20@mails.jlu.edu.cn}{junwei20},
			\href{mailto:gongxy20@mails.jlu.edu.cn}{gongxy20}\}@mails.jlu.edu.cn,
			\href{mailto:lus@jlu.edu.cn}{lus@jlu.edu.cn}}}
	
	\maketitle
	
	\begin{abstract}
		Generative Adversarial Networks (GAN) is an adversarial model, and it has been demonstrated to be effective for various generative tasks. However, GAN and its variants also suffer from many training problems, such as mode collapse and gradient vanish. In this paper, we firstly propose a general crossover operator, which can be widely applied to GANs using evolutionary strategies. Then we design an evolutionary GAN framework named C-GAN based on it. And we combine the crossover operator with evolutionary generative adversarial networks (E-GAN) to implement the evolutionary generative adversarial networks with crossover (CE-GAN). Under the premise that a variety of loss functions are used as mutation operators to generate mutation individuals, we evaluate the generated samples and allow the mutation individuals to learn experiences from the output in a knowledge distillation manner, imitating the best output outcome, resulting in better offspring. Then, we greedily select the best offspring as parents for subsequent training using discriminator as an evaluator. Experiments on real datasets demonstrate the effectiveness of CE-GAN and show that our method is competitive in terms of generated images quality and time efficiency.
	\end{abstract}
	
	
	\section{Introduction}
	Generative adversarial networks (GAN) \cite{b1} is a deep learning model and one of the most promising methods for unsupervised learning on complex distributions in recent years. The framework produces quite good output through mutual game learning of (at least) two modules in the framework: generative model and discriminative model, \textit{i.e.}, generator and discriminator. Deep neural networks are generally used as generator or discriminator in practice \cite{b11}.
	
	During the training process, the goal of the generator is to generate samples that are very similar to the real samples in order to deceive the discriminator. Adversarially, the goal of the discriminator is to distinguish as much as possible between the samples generated by the generator and the real samples. Generator and discriminator form a dynamic convex-concave game, which is the basic idea of GAN. 
	
	However, its training is challenging, and problems such as mode collapse and gradient vanish are common \cite{b13}. These problems usually stem from the imbalance between the discriminator and generator \cite{b40}. 
	
	Evolutionary generative adversarial networks (E-GAN) introduces the idea of evolutionary computation \cite{b20}. It maintains a generator population that generates individuals with different loss functions as mutation operators and promoted by evaluating individuals in terms of both quality and diversity, thereby benefiting from the multiple loss functions. Cooperative dual evolution based generative adversarial networks (CDE-GAN) expands the concept of population to discriminator and uses soft mechanism to connect the two populations \cite{b21}. Mu \textit{et al.} define different mutation operator, \textit{i.e.}, a distribution indicating realness \cite{b22}. Multi-objective evolutionary generative adversarial networks (MO-EGAN) considers quality and diversity as conflicting objectives, and defines the evaluation of generators as a multi-objective problem \cite{b23}. Mustangs is inspired by E-GAN and applies different loss functions to competitive co-evolutionary algorithm \cite{b24}. In addition, some researchers combine GANs with evolutionary strategies in their own ways \cite{b10,b25,b28,b29,b30}. Garciarena \textit{et al.} use crossover in \cite{b30}, which is as common in biology as mutation. But the crossover operator it used is deeply coupled with itself and cannot be directly used in other methods. 
	
	Knowledge distillation is a model compression method, also a training method based on the “teacher-student network”, in which the knowledge contained in a trained model is distilled and extracted into target model. The target model is often a smaller model that is compressed and has better generalizability. Knowledge distillation as a means of compressing networks has been used for GAN \cite{b32,b33}. Proximal distilled evolutionary reinforcement learning (PDERL) \cite{b34} utilizes this technique to optimize evolutionary reinforcement learning (ERL) \cite{b35}.
	
	In this paper, we propose a crossover operator utilizing knowledge distillation that can be easily ported to any GAN using evolutionary computation, regardless of whether its encoding genes are weights or architecture.
	This paper makes the following contributions:
	
	\begin{itemize}
		\item We propose a universal backpropagation-based cross genetic operator.
		\item We propose a framework called C-GAN that uses the crossover operator, which can integrate GANs with evolutionary computation regardless of what genes encoded. 
		\item We integrate the crossover operator as part of an algorithm called evolutionary generative adversarial networks with crossover (CE-GAN), which unifies the evolutionary and learning processes. 
		\item Experiments show CE-GAN performs better on the CIFAR-10 dataset. The combination of crossover operator and E-GAN is able to obtain competitive results with less time cost.
	\end{itemize}

	\section{Related Works}
	
	\subsection{Evolutionary Generative Adversarial Networks}
	E-GAN designs an adversarial framework between a discriminator and a population of generators \cite{b20}. Specifically, it assumes that the generators no longer exist as individuals, but in the form of populations to confront with discriminator. From an evolutionary point of view, the discriminator can be considered as a changing environment during the evolutionary process. 
	
	During each evolutionary step in \cite{b20}, the evolution of the generator consists of three steps: variation, evaluation and selection. The variation includes only mutation operation, and different loss functions are chosen as mutation operators to obtain different offspring. The new offspring obtained after mutation needs to be evaluated for their generative performance and quantified as the corresponding fitness. We will cite the fitness function in \cite{b20} as $F_{E-GAN}$ in this paper to avoid confusion. 
	
	After measuring the generation performance of all offspring, according to the principle of survival of the fittest, the updated generators are selected as the parents for a new round of training.

	\subsection{Variation operators}
	In the evolutionary computational community, variation often consists of two components: mutation and crossover. The point of crossover is that the best behavior of two parents can be combined. But GANs combined evolutionary computation often do not have crossover operators.
	
	The backpropagation-based crossover operator named $Q$-filtered distillation crossover is proposed in PDERL \cite{b34}. It is experimentally demonstrated that in reinforcement learning (RL), the operator can ensure that offspring inherit the behavior of their parents.

	\subsection{Gradient Penalty}
	Gradient penalty (GP), proposed in \cite{b17}, is a method to achieve the Lipschitz constraint required in \cite{b16}. It can be used to regularize the discriminator to support updating the generator.
	
	Although it was originally used to improve Wasserstein GAN (WGAN) \cite{b16}, Wang \textit{et al.} find that E-GAN and GP term are orthogonal \cite{b20}.

	\section{Preliminaries}

	\subsection{Notation}
	We combine the symbols used in \cite{b36,b20}. Most GANs can be formally defined in the following form:
	\begin{equation}
		L_{D}^{GAN}=\mathbb{E}_{x \sim p_{data}}[f_{1}(C(x))]+\mathbb{E}_{z \sim p_{z}}[f_{2}(C(G(z)))]
	\end{equation}
	\begin{equation}
		L_{G}^{GAN}=\mathbb{E}_{x \sim p_{data}}[g_{1}(C(x))]+\mathbb{E}_{z \sim p_{z}}[g_{2}(C(G(z)))]
	\end{equation}
	where $f_1$, $f_2$, $g_1$, $g_2$ are scalar-to-scalar functions, and the GAN and its variants with different loss functions correspond to these functions differently. $p_{data}$ is the distribution of real data, $p_z$ is the distribution of sampling noise (usually a uniform or normal distribution). $G(z)$ is the output of the generative network $G$, and its distribution $p_g$ is expected to be fitted into $p_{data}$. $C(x)$ is the non-transformed discriminator output. In the original GAN, $D(x)=sigmoid(C(x))$, where $D(x)$ is the output of the discriminative network $D$. In most GANs, $C(x)$ can be interpreted as how realistic the input data is \cite{b36}.

	\subsection{Discriminator objective function}\label{3-2}
	In the original GAN \cite{b1}, discriminator objective function is defined as the following form:
		\begin{equation}
		L_{D}=-\mathbb{E}_{x \sim p_{data}}[\log D(x)]-\mathbb{E}_{y \sim p_{g}}[\log (1-D(y))]\label{obj:discriminator}
	\end{equation}
	
	\subsection{Generator objective function}\label{3-3}
	E-GAN uses three different minimization objective functions as mutation operators.
	
	\begin{itemize}
	\item Minimax mutation in original GAN \cite{b1}:
		\begin{equation}
			M_{G}^{minimax}=\frac{1}{2} \mathbb{E}_{z \sim p_{z}}[\log (1-D(G(z)))]
		\end{equation}
	\item Heuristic mutation in non-saturated GAN (NS-GAN) \cite{b13}:
		\begin{equation}
			M_{G}^{heuristic}=-\frac{1}{2} \mathbb{E}_{z \sim p_{z}}[\log (D(G(z)))]
		\end{equation}
	\item Least-squares mutation in least squares GAN (LSGAN) \cite{b18}:
		\begin{equation}
			M_{G}^{least-square}=\mathbb{E}_{z \sim p_{z}}[(D(G(z))-1)^{2}]
		\end{equation}
	\end{itemize}

	\subsection{Fitness function}
	An evaluation criterion to measure the quality of individuals is needed in the evolutionary algorithm. The different schemes utilize different fitness, such as Inverted Generational Distance (IGD) \cite{b30}, GAN objective \cite{b24,b25}, Fréchet Inception Distance (FID) \cite{b24,b25}, Koncept512 \cite{b29}, $F_{E-GAN}$ and its variants \cite{b20,b21,b22,b23}, and a mixture of them.
	
	Fitness function $F_{E-GAN}$ \cite{b20} consists of quality fitness $F_q$ and diversity fitness $F_d$:
		
	\begin{equation}
		F_{E-GAN}=F_q+\gamma F_d\label{obj:E-Fitness}
	\end{equation}

	\begin{equation}
	F_q=\mathbb{E}_z[D(G(z))]
	\end{equation}
	
	\begin{equation}
	F_{d}=-\log \|\nabla_{D}-\mathbb{E}_{x}[\log D(x)]-\mathbb{E}_{z}[\log (1-D(G(z)))]\|
	\end{equation}
	where $\gamma>0$ balances two measurements.

	\section{Method}
	
	\begin{figure*}[tbp]
		\centering
		\includegraphics[scale=0.70]{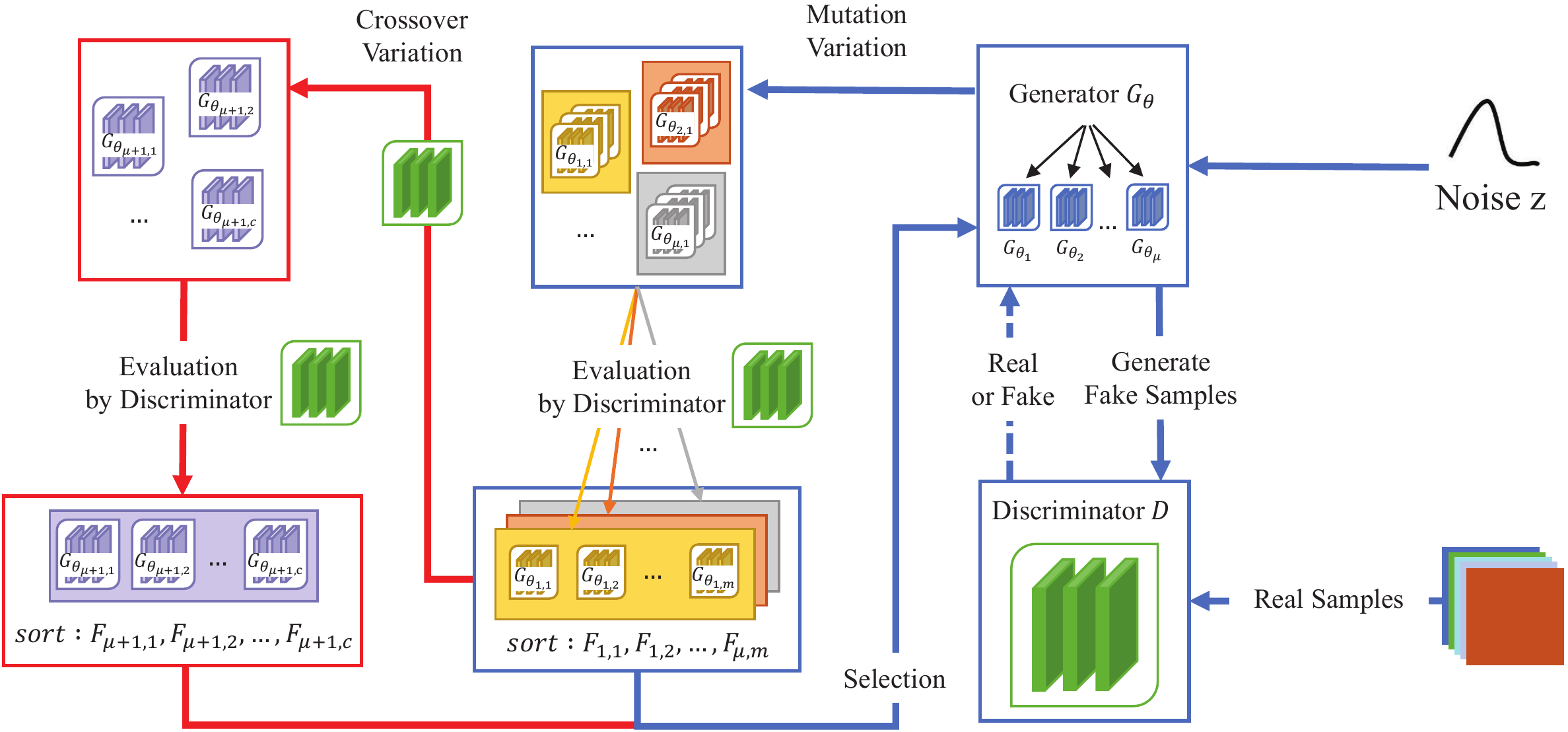}
		\caption{The proposed C-GAN framework. C-GAN adds crossover variation to leverage the experience of mutant individuals.}
		\label{fig1}
	\end{figure*}
	
	\subsection{C-GAN framework}
	The framework C-GAN designed in this paper evolves a population of generators $G$ in a given dynamic environment (discriminator $D$). Each individual in the population represents a possible solution in the parameter space \cite{b20}. C-GAN framework is shown in Fig.~\ref{fig1}. 
	
	In each evolutionary step, individuals $\{G_{\theta_1}, \cdots, G_{\theta_\mu}\}$ in the population $G_\theta$ act as parents to generate offspring $\{G_{\theta_{1,1}}, G_{\theta_{1,2}}, \cdots, G_{\theta_{\mu,m}}\}$ by mutation operators. After evaluating them using fitness function $F(\cdot)$, the environmental fitness of mutation individuals is obtained. Then crossover parents are selected according to these fitness values, and crossover individuals $\{G_{\theta_{\mu+1,1}}, G_{\theta_{\mu+1,2}}, \cdots, G_{\theta_{\mu+1,c}}\}$ are got by crossover operator. Evaluating the fitness of crossover individuals. All offspring are selected based on their fitness values, and the best part survives and evolves into the next generation as parents.

	After evolutionary step, just like the usual GAN, the discriminator $D$ is updated. Thus the environment is also dynamically changing as the population evolves and is able to provide continuous evolutionary pressure on the population. The objective function of $D$ is chosen from the discriminator loss function of the original GAN.
	
	
	\subsection{Crossover}
	In this subsection, we present the $C$-filtered knowledge distillation crossover \textit{w.r.t.} the $Q$-filtered behavior distillation crossover \cite{b34}. Due to the difference between GAN and RL, our crossover operator differs from it in terms of sample filtering and basis network selection. 
	Considering the generality of the operator, we choose to perform pixel-level imitation learning from the output so that it can be applied to a more general framework with evolutionary strategies. 
	
	The crossover operation of a pair of parent networks is as follows:
	\begin{itemize}
	\item One of the parent models is used as the basis for the offspring model, and the generated results of the parents with the same inputs are used as experiences for the offspring to imitate and learn. 
	\item The process transfers the knowledge of parents to offspring by means of knowledge distillation. 
	\end{itemize}
	
	The outputs of parents for the same inputs are likely to be completely different. So a natural question is that how should the offspring imitate both networks at the same time? The solution is that the child chooses the better output to imitate for each input. The definition of good and bad output is left to the discriminator, \textit{i.e.}, $C(x)$. The higher the $C(x)$, the more worthy of study. 
	
	The formal representation of the $C$-filtered knowledge distillation loss used to train the offspring is as follows:
	\begin{equation}
		\begin{split}
			L_{C}&=\sum_{i}^{n}\|G_{child}(z_{i})-G_{x}(z_{i})\|^{2} \mathbb{I}_{C(G_{x}(z_{i}))>C(G_{y}(z_{i}))} \\
			&+\sum_{j}^{n}\|G_{child}(z_{j})-G_{y}(z_{j})\|^{2} \mathbb{I}_{C(G_{y}(z_{j}))>C(G_{x}(z_{j}))}
		\end{split}
	\end{equation}
	where $\sum$ is the sum of samples generated by a batch of size \textit{n} noise inputs from two parent individuals. $G_x$ and $G_y$ represent the generative strategies of parents, and $G_{child}$ represents the one of child. The parameters of child network are initialized to the one with higher fitness among the parents. The offspring network learns the knowledge of parents by minimizing this function $L_C$ during training. The whole procedure of the crossover operator is summarized in Algorithm \ref{alg1}. 
	
	\begin{algorithm}[tbp]
		\caption{$C$-filtered Knowledge Distillation Crossover. }
		\label{alg1}
		\SetKwInOut{Input}{Require}
		\SetKwInOut{Output}{Output}
		\Input {The parameters of parents $\theta_x$, $\theta_y$. Crossover batch size $n$. Adam hyper-parameters $\alpha$, $\beta_1$, $\beta_2$.}
		Initialize the parameters of child $\theta_0$ with the parameters of better parent\\
		Sample a batch of $\{{z_i}\}_{i=1}^n \sim p_z$\\
		$g_\theta\gets\nabla_{\theta_0}L_C(\{z_i\}_{i=1}^n,\theta_0)$\\
		$\theta_{child}\gets Adam(g_\theta,\theta_0, \alpha, \beta_1, \beta_2)$\\
	\end{algorithm}
	
	There is still a problem: how to select parents from the multiple networks for crossover? We used the greedy strategy score function $W$ defined in \cite{b34}:
	\begin{equation}
		W=F(G_x)+F(G_y)
	\end{equation}
	
	
	
	\subsection{Evaluation}
	
	The fitness $F$ we use is given by the discriminator forward propagation only:
	\begin{equation}
		F_{C-GAN}=\mathbb{E}_z[C(G(z))]\label{obj:CE-Fitness}
	\end{equation}
	
	\subsection{Selection}
	There are two questions on the selection range of the next generation parents: whether the mutation individuals are still candidates? and whether the current parents are candidates?
	
	For the former question, we have experimentally demonstrated that only crossover offspring candidates do not lead to significant gains, and doing so does not reduce the computational complexity, the parent selection of the crossover operator needs to be evaluated first for the mutation offspring. Therefore, mutant offspring should not be excluded from selection.
	
	As for the latter issue, Existing studies re-evaluate current parents using updated discriminators and include them as next generation parent candidates \cite{b21,b23}. Experiments on our framework show that this strategy enhances training stability. But it will slow down convergence, drag down the final result and increase computational effort. So we will not consider transferring the current parents to the next generation.

	\subsection{CE-GAN algorithm}
	The crossover operator improves population stability, drives the possible solutions to a more adaptive parameter space, and the low-cost fitness function reduces time cost. The framework implemented in this paper is based on E-GAN, named CE-GAN. The complete CE-GAN training procedure is summarized in Algorithm \ref{alg2}. 
	
	The use of fitness function \eqref{obj:CE-Fitness} can effectively alleviate gradient vanish caused by $F_{E-GAN}$. The behavior of crossover individuals to learn multiple generators at the same time can also avoid mode collapse to some extent.

	\begin{algorithm}[tbp]
		\caption{Evolutionary Generative Adversarial Networks with Crossover (CE-GAN).}
		\label{alg2}
		\SetKwInOut{Input}{Require}
		\SetKwInOut{Output}{Output}
		\Input {Mutation batch size $m$. The updating steps of discriminator per iteration $n_D$. The number of parents $\mu$. The number of mutations $n_m$. The number of crossovers $n_c$. Adam hyper-parameters $\alpha$, $\beta_1$, $\beta_2$.}
		Initial the parameters of discriminator $\omega_0$ and generators $\{\theta_0^1,\theta_0^2,\cdots,\theta_0^\mu\}$\\
		\ForEach{training iteration} {
			\For{$k=1,\cdots,n_D$} {
				Sample a batch of $\{{x_i}\}_{i=1}^m \sim p_{data}$, and a batch of $\{{z_i}\}_{i=1}^m \sim p_z$\\
				Updating $g_{\omega}$:
				\begin{equation}
				\setlength\abovedisplayskip{-6pt}
				\setlength\belowdisplayskip{2pt}
				\nonumber
				\begin{split}
				g_{\omega} &\leftarrow \nabla_{\omega}[\frac{1}{m} \sum_{i=1}^{m} \log D_{\omega}(x_i) \\
				&+ \frac{1}{m} \sum_{j=1}^{\mu} \sum_{i=1}^{m / \mu} \log (1-D_{\omega}(G_{\theta^{j}}(z_i)))]
				\end{split}
				\end{equation}\\
				$\omega\gets Adam(g_\omega,\omega, \alpha,\beta_1,\beta_2)$\\
			}
			\For{$j=1,\cdots,\mu$} {
				\For{$h=1,\cdots,n_m$} {
					Sample a batch of $\{{z_i}\}_{i=1}^m \sim p_z$\\
					$g_{\theta^{j,h}}\gets\nabla_{\theta^j}M_G^h(\{z_i\}_{i=1}^m,\theta^j)$\\
					$\theta_{child}^{j,h}\gets Adam(g_{\theta^{j,h}},\theta^j, \alpha,\beta_1,\beta_2)$\\
					Caculate $F^{j,h}$\\
				}
			}
			\For{$p=1,\cdots,\mu\times n_m$} {
				\For{$q=p+1,\cdots,\mu\times n_m$} {
					$W^{p,q}{\gets F}^{j_p,h_p}+F^{j_q,h_q}$\\
				}
			}
			$\{W^{p_1,q_1},W^{p_2,q_2},\cdots\}\gets sort(\{W^{p,q}\})$\\
			\For{$r=1,\cdots,n_c$} {
				$\theta_{child}^r\gets crossover(\theta_{child}^{j_{p_r},h_{p_r}},\theta_{child}^{j_{q_r},h_{q_r}})$\\
				Caculate $F^r$\\
			}
			$\{F^1,F^2,\cdots\}\gets sort(\{F^{j,h}\}\bigcup\{F^r\})$\\
			\For{$s=1,\cdots,\mu$}{
			    $\theta^s\gets correspond(F^s)$
			}
		}
	\end{algorithm}

\section{Experiments}

\subsection{Experimental configuration}
The experiments were conducted on the real image dataset CIFAR-10  \cite{b43}. CIFAR-10 is a dataset for identifying universal objects, containing 10 categories of RGB color images about real objects in the real world. The images in it are not only noisy, but also the proportions and features of the objects are not the same, which brings great difficulties to recognition. The size of the images is $32\!\times\!32$.


The baselines used for comparison including GAN-Minimax, GAN-Least-square, and GAN-Heuristic which represent the methods using respectively the corresponding single mutation operator, demonstrating that our approach benefits primarily from the framework rather than single improvement in network structure for a particular loss function. In addition, we will compare with E-GAN \cite{b20} to demonstrate the effectiveness of the proposed crossover operator.

We use the inception score \cite{b44} with Fréchet Inception Distance (FID) \cite{b45} to quantitatively evaluate the generative performance. They are the two most frequently used metrics in researches regarding GANs, and FID is considered to outperform other metrics, including inception score \cite{b40}. The higher inception score and the lower FID, the better the quality of the generated images. We randomly generate 50k images to calculate inception score and FID. Without special instructions, the experiments in this paper will be trained for 100k generations.

The experiments were trained on a single NVIDIA TITAN Xp GPU with 12GB memory and Intel Xeon E5-2620 v2 CPU. Our code is publicly available\footnote{\url{https://github.com/AlephZr/CE-GAN}}.

\subsection{Hyperparameters analysis}
We have tried to use the least squares loss function \cite{b18} as discriminator objective function just like in \cite{b22}, but the experimental results did not show any significant advantage. Therefore, we use the same loss function of the original GAN as E-GAN.

The architecture of the generators and discriminator networks are the same as E-GAN \cite{b20}, which are fine-tuned DCGAN networks \cite{b11}. The values of hyperparameters shared with E-GAN are the same, except for the learning rate $\alpha$. The introduction of crossover operator accelerates convergence, and a lower learning rate helps converge to better results. For CE-GAN specific hyperparameters, we set the cross batch size $n$ to 256, which is consistent with the number of samples used when calculating fitness. In this way, the samples generated when evaluating the fitness of mutation individuals can be reused during crossover variation.

There are two more hyperparameters that are closely related to the performance of CE-GAN experiments, namely the choice of fitness function and the crossover population size $n_c$. The fitness function is directly related to the choice of generator and therefore affects the performance of CE-GAN. Also it requires additional computation and the time cost cannot be neglected when the training lasts for a large number of generations. A larger population in evolutionary computation often leads to greater time consumption while bringing more advantages. Therefore we should analyze the effect of the number of crossover populations on the results.

In order to choose the appropriate hyperparameters for CE-GAN, we conducted experiments on CIFAR-10 to compare the inception score, as well as wall-clock times (interference from the calculation of inception score has been excluded).

\begin{figure}[tbp]
	\centerline{\includegraphics[scale=0.58]{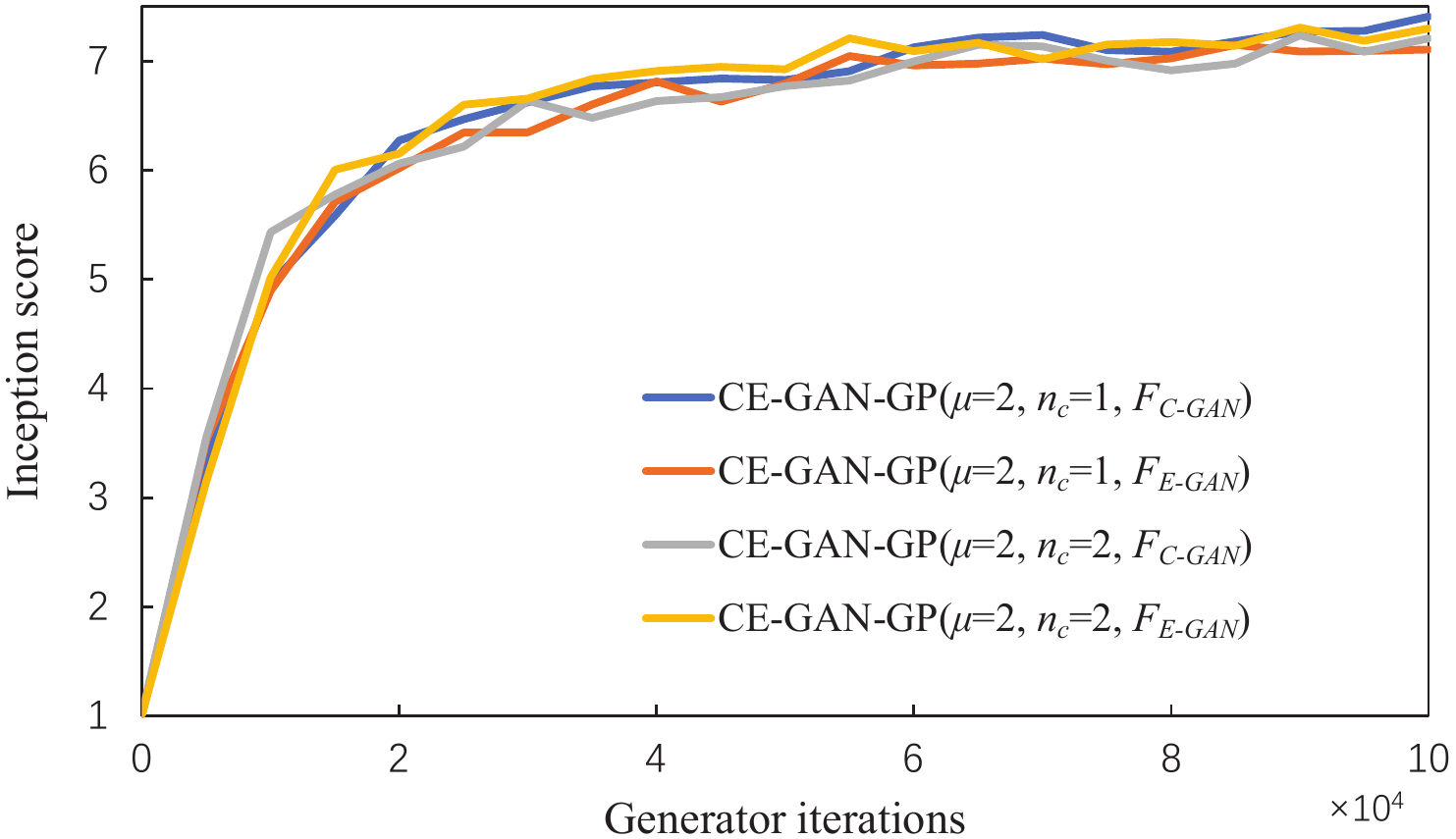}}
	\caption{Inception score evaluation of different CE-GANs with various crossover size $n_c=\{1,2\}$ and fitness score function $F=\{F_{E-GAN},F_{C-GAN}\}$.}
	\label{fig2}
\end{figure}

As shown in Fig.~\ref{fig2}, with population number set to 2, we illustrate inception score of CE-GAN using two different functions and crossover populations. The objects marked with “-GP” here and after indicate that GP term is used. CE-GAN with $n_c=1$ and $F_{C - GAN}$ has the best generation performance. Interestingly, the two fitness functions perform different strengths \textit{w.r.t.} different numbers of crossover population $n_c$. When $n_c=1$, CE-GAN using $F_{C - GAN}$ is significantly better than that using $F_{E - GAN}$. Throughout most of the training process, the fold representing CE-GAN with $F_{C - GAN}$ is higher than the one representing CE-GAN with $F_{E - GAN}$. When $n_c=2$, the situation is reversed, but the difference between the two is not as obvious as the previous group. It can be seen that $F_{C - GAN}$ declines with the increase of crossover population, while $F_{E - GAN}$ is the opposite. Experiments with different crossing populations repeated this phenomenon. We believe that this is because the crossover individual itself is good enough, so greedy selection can get a good result when the number of crossover is 1. The $F_d$ contained in $F_{E - GAN}$ is not accurate enough, but will have a counterproductive effect at this time. As the population increases, the side effect of $F_{C - GAN}$ evaluating individuals based only on quality begins to become prominent, and the diversity of generated images declines, which instead lowers inception score. Diversity measure of $F_{E - GAN}$ plays its role. More generator networks mean more candidates, which improves generation performance.

\begin{figure}[tbp]
	\centerline{\includegraphics[scale=0.46]{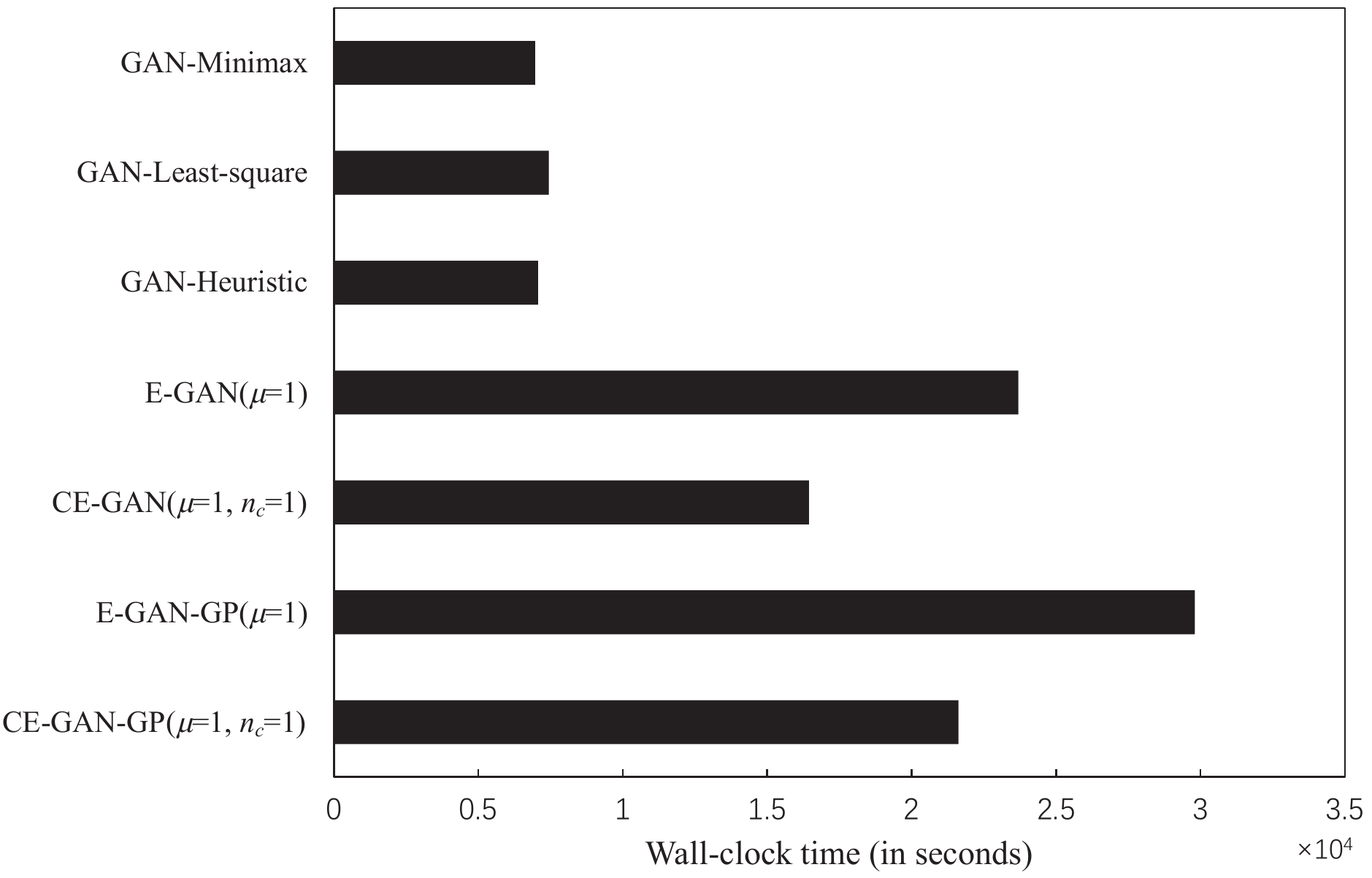}}
	\caption{Wall-clock time of training different methods 100k iterations on the CIFAR-10 dataset.}
	\label{fig3}
\end{figure}

Fig.~\ref{fig3} shows the wall-clock time of training different methods 100k iterations. The GAN-Minimax, GAN-Least-square, and GAN-Heuristic maintain only one generator network, so the time cost for training is less and similar. CE-GAN with $F_{C - GAN}$ costs just over twice as much time as them while maintaining four generative networks. Even with GP term added, the time cost is about three times that of them. The time cost of E-GAN is significantly higher, which is mainly due to the extra calculation of the discriminator gradient information by $F_d$ in $F_{E - GAN}$. With the increase in the number of parents $\mu$, the time required to calculate $F_d$ becomes more and more difficult to ignore.

We also conduct ablation experiments on the crossover batch size $n$, which show that the final result is positively correlated with $n$ within a certain range. Therefore, the crossover batch size is set to 256 in this paper, which is consistent with the number of samples used to evaluate individuals, so as to fully reuse the generated samples. 

In summary, comprehensively generative performance and wall-clock time considerations, for subsequent experiments, we use $F_{C - GAN}$ as well as one crossover population for training. Hyperparameters default values $\alpha=1e\!-\!4$, $\beta_1=0.5$, $\beta_2=0.999$, $n_D=3$, $n_m=3$, $n_c=1$, $m=32$, $n=256$.

\subsection{Crossover evaluation}
We considered using a common biologically-inspired crossover operator, but experiments show that this is not a good idea because it did not work as expected when it is applied to GANs, and the offspring troubled by catastrophic forgetting while having a high time cost. Therefore, we propose $C$-filtered knowledge distillation crossover by referring to $Q$-filtered knowledge distillation crossover in PDERL \cite{b34}.

PDERL implementation\footnote{\url{https://github.com/crisbodnar/pderl}} chosen a less capable network as the initial child, but we choose the better one, which arises from the difference between GAN and RL. The networks of GAN tend to be deeper, the generated samples are more time-sensitive, and the sample batch size selected for efficiency are smaller.

\begin{figure}[tbp]
	\centerline{\includegraphics[scale=0.305]{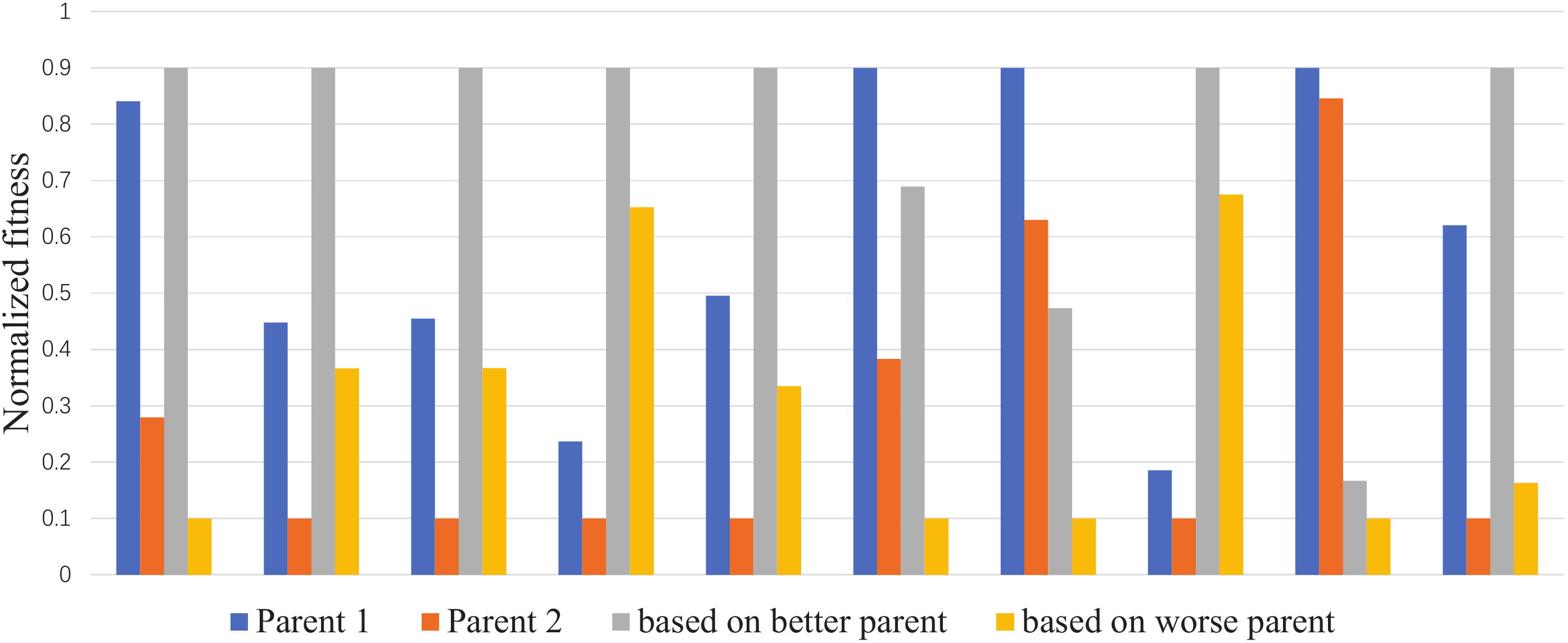}}
	\caption{Normalised crossover performance.}
	\label{fig4}
\end{figure}

To demonstrate it, we experimentally compared the two crossover operators, \textit{i.e.}, based on better parent and based on worse parent. The fitness of offspring is a good indicator to measure quality of crossover operator. Fig.~\ref{fig4} plots ten randomly selected parent pairs, with each set of bars including the fitness of two parents along with two types of crossover offspring. These values were normalized to [0.1, 0.9]. Seven of the ten sets performed best with crossover individual based on better parent. Not once did the crossover individual based on worse parent have higher fitness than that based on better parent, and only twice did it have higher fitness than the parents. Offspring based on worse parent usually perform lower than better parent, or even lower than worse parent. We believe this is because a worse initial network requires more samples to learn knowledge, which is uneconomical for efficiency reasons.

\begin{figure}[tbp]
	\centerline{\includegraphics[scale=0.473]{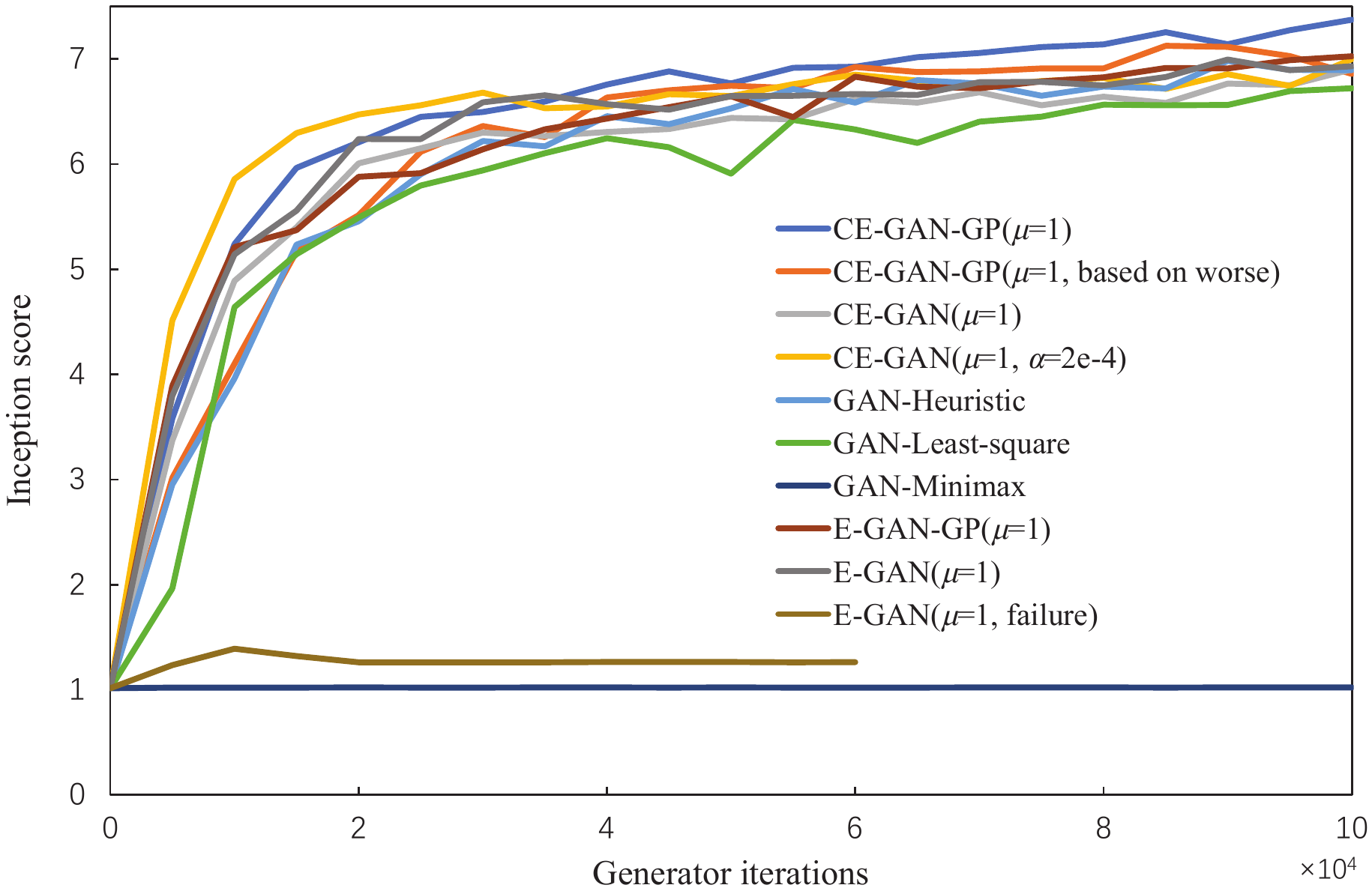}}
	\caption{Inception score of different methods on CIFAR-10 over generator iteration. $\mu$ is the number of parents, $\alpha$ is the Adam optimizer learning rate.}
	\label{fig5}
\end{figure}

Fig.~\ref{fig5} is consistent with expectations in the actual training. The fold representing better parental crossover operators outperforms the fold representing worse one. The framework with crossover operator based on the worse parent  converges more slowly and is more unstable.

\begin{figure}[tbp]
	\centerline{\includegraphics[scale=0.405]{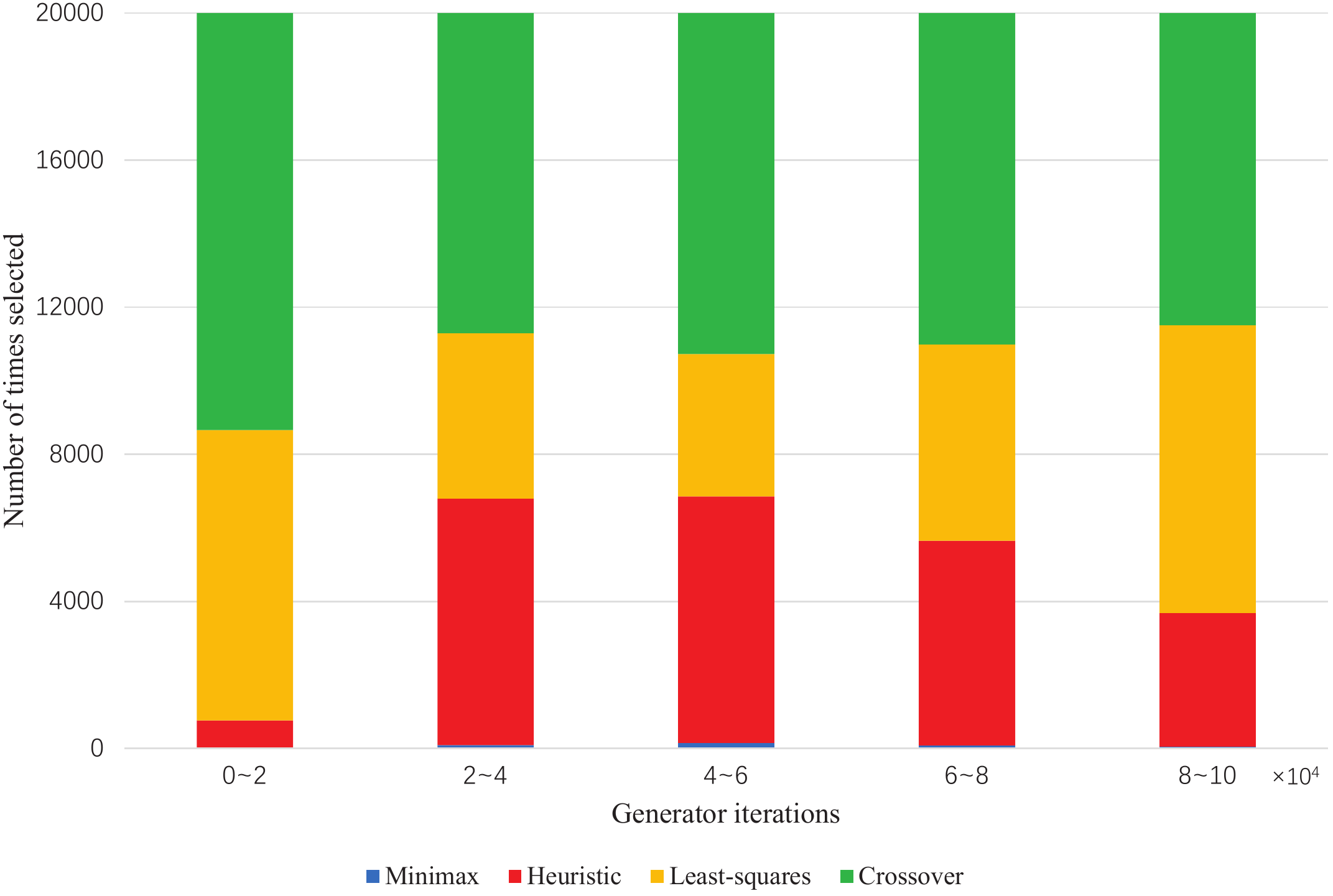}}
	\caption{Stacked bar chart of the selection of operators in the CE-GAN training process.}
	\label{fig6}
\end{figure}

In addition, Fig.~\ref{fig6} records the selection of operators for CE-GAN during the training process. From this, the competitiveness of different operators in different phases can be analyzed. Crossover operator occupies a great weight throughout the training process. The most affected one is minimax mutation operator, which will hardly be selected anymore, but this does not mean that it is meaningless. The selection here only means that it is selected as the next generation parent, and more often than not, the unselected mutation network migrates knowledge to the crossover network. This is one of the advantages of crossover operator, and it can take advantage of poorly performing generative networks, rather than simply discarding them.

\subsection{Generative performance evaluation}
To illustrate the superiority of proposed CE-GAN over baselines, Fig.~\ref{fig5} plots inception score for training process under the same network architecture. 
\begin{itemize}
	\item As a single mutation operator network, GAN-Minimax fails to train, and GAN-Least-square is unstable. Although GAN-Heuristic performs better, its convergence is slower compared to the evolutionary GAN frameworks, and the generation performance is inferior.
	\item CE-GAN without GP term has similar performance to E-GAN on inception score within 100k iterations, but it converges more slowly. This is because CE-GAN uses a smaller learning rate $\alpha$. CE-GAN with the same learning rate of $2e\!-\!4$ as E-GAN has the fastest convergence speed of all methods and a generative performance that is not weaker than E-GAN.
	\item CE-GAN with GP term converges the second fastest, even though the learning rate $\alpha$ is smaller than it in E-GAN. This is due to the improvement of GP term for the discriminator. The discriminator participates in the crossover and dominates the evaluation. Crossover individuals tend to outperform their parents, and the evolutionary framework can select the best from offspring, both of which work together to accelerate the learning process. There is no similar acceleration effect in E-GAN.
	\item The generation performance of CE-GAN with GP term is also the best, and even at 100k iterations, there is still a clear upward trend. A better discriminator can generate better individuals, especially crossover individuals, and can evaluate individuals more accurately.
	\item Fig.~\ref{fig5} also records a failed training of E-GAN with $F_{E-GAN}$. As mentioned earlier, the training process stalled due to the unreasonably high scores assigned by $F_{E-GAN}$ to the gradient-vanishing minimax mutation operator. Whereas no such situation occurred in numerous experiments of CE-GAN, indicating that our method can indeed effectively mitigate gradient vanish.
\end{itemize}

The increase in the number of maintained populations is significant for CE-GAN. The inception score and FID for different populations of E-GAN and CE-GAN are listed in Table~\ref{tab1}, where the data marked $\S$ is quoted from \cite{b20}. We use experimental setup from the literature  \cite{b20} with codes provided by authors\footnote{\url{https://github.com/WANG-Chaoyue/EvolutionaryGAN-pytorch}} to implement the results of E-GAN. 
\begin{itemize}
	\item When $\mu=1$ and no GP term, inception score of CE-GAN is slightly lower than E-GAN, but its FID is better than E-GAN, and even better than E-GAN with GP term.
	\item When GP term is not used, CE-GAN with a larger learning rate performs better, no matter which metric to adopted.
	\item CE-GAN outperforms 24 generative networks ($\mu = 8$) of E-GAN using just 4 generative networks ($\mu = 1$), regardless of FID or inception score. 
	\item CE-GAN also performs better in terms of inception score when the population is larger.
\end{itemize}

\begin{table}[tbp]
	\caption{Comparison With E-GAN On Cifar-10.}
	\begin{center}
		\begin{tabular}{llcc}
			\hline
			\rule{0pt}{10pt} & \makecell[c]{\textbf{Methods}} & {Inception score} & {FID} \\
			\hline
			\rule{0pt}{10pt} & {E-GAN ($\mu = 1$)} & {$6.93 \pm 0.10$} & {$36.2$$^\S$} \\ 
			\rule{0pt}{10pt} & {E-GAN-GP ($\mu = 1$)} & {$7.03 \pm 0.06$} & {$33.2$$^\S$} \\
			\rule{0pt}{10pt} & {E-GAN-GP ($\mu = 2$) $^\S$} & {$7.23 \pm 0.08$} & {$31.6$} \\ 
			\rule{0pt}{10pt} & {E-GAN-GP ($\mu = 4$) $^\S$} & {$7.32 \pm 0.09$} & {$29.8$} \\ 
			\rule{0pt}{10pt} & {E-GAN-GP ($\mu = 8$) $^\S$} & {$7.34 \pm 0.07$} & {$27.3$} \\ 
			\hline
			\rule{0pt}{10pt} & {CE-GAN ($\mu = 1$)} & {$6.90 \pm 0.08$} & {$32.4$} \\ 
			\rule{0pt}{10pt} & {CE-GAN ($\mu = 1,\alpha = 2e\!-\!4$)} & {$7.00 \pm 0.04$} & {$29.7$} \\ 
			\rule{0pt}{10pt} & {CE-GAN-GP ($\mu = 1$)} & {$7.38 \pm 0.11$} & \textbf{26.4} \\ 
			\rule{0pt}{10pt} & {CE-GAN-GP ($\mu = 2$)} & \textbf{7.41 $\pm$ 0.05} & {$26.6$} \\ 
			\hline
		\end{tabular}
		\label{tab1}
	\end{center}
\end{table}

\begin{figure}[tbp]
	\centerline{\includegraphics[scale=0.9]{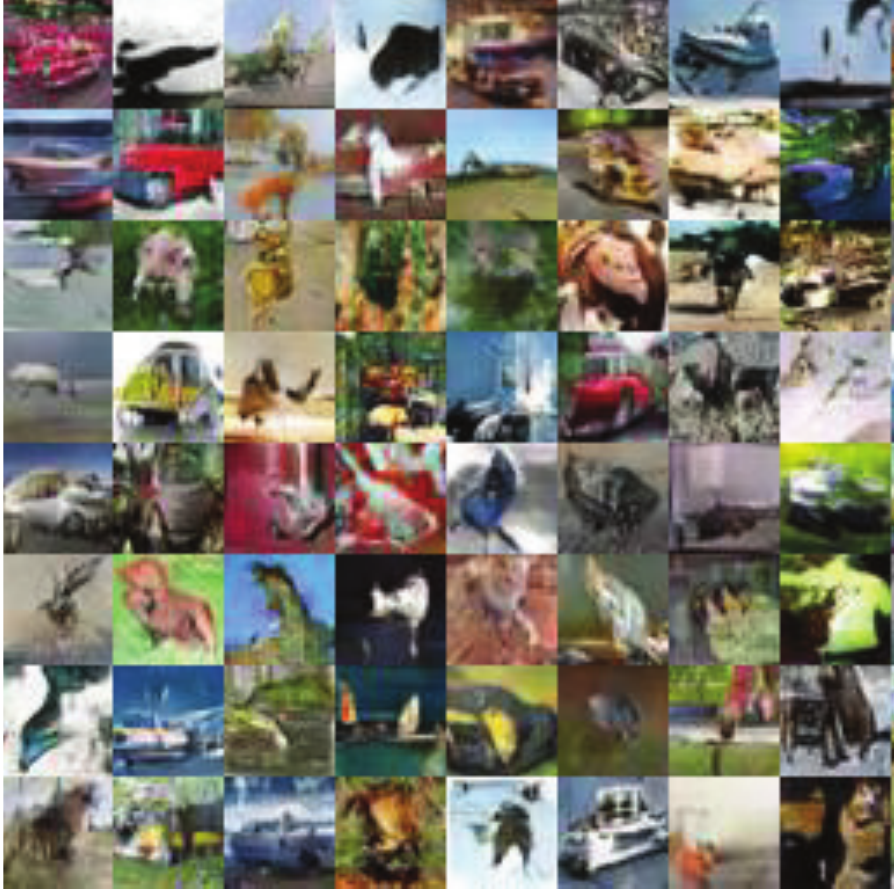}}
	\caption{Images generated by CE-GAN ($\mu=1$) on CIFAR-10 dataset.}
	\label{fig7}
\end{figure}

The generated images of CE-GAN are shown in Fig.~\ref{fig7}. It can be seen that our method does not suffer from mode collapse.

\section{Conclusion}
In this paper, we propose a crossover operator that can be widely applied to evolutionary GANs. We design a framework C-GAN that combines crossover operators and GANs with evolutionary computation. We combine the crossover operator with E-GAN to implement CE-GAN that can take advantage of generative networks. We experimentally demonstrate that our approach has better generation performance with less time cost.

Our future work is as follows:
\begin{itemize}
	\item The final performance of the generator decreases instead when trying to provide more networks for the crossover operator to imitate. We will explore the intrinsic reasons for this phenomenon in future work. 
	\item We will also try to answer some questions that arise from the experiments, such as why the least squares loss function does not bring advantages, and hope to theoretically prove and infer the experimental principles, as well as to extend the work in this paper to more and more complex datasets.
\end{itemize}

\section*{Acknowledgment}

This work was supported by the Natural Science Research Foundation of Jilin Province of China under Grant No. 20180101053JC; the National Key R\&D Program of China under Grant No. 2017YFB1003103; and the National Natural Science Foundation of China under Grant No. 61300049.

\end{document}